\def\paperID{9434} 
\def\confName{CVPR}
\def\confYear{2024}

\def\paperTitle{SAVE: Protagonist Diversification with \underline{S}tructure \underline{A}gnostic \underline{V}ideo \underline{E}diting}

\def\authorBlock{Yeji Song\textsuperscript{1}\hspace{.2cm}
Wonsik Shin\textsuperscript{1}\hspace{.2cm}
Junsoo Lee\textsuperscript{2}\hspace{.2cm}
Jeesoo Kim\textsuperscript{2}\hspace{.2cm}
Nojun Kwak\textsuperscript{1}\hspace{.2cm} \\
\textsuperscript{1}{Seoul National University}\hspace{.2cm}
\textsuperscript{2}{NAVER Webtoon AI} \\
{\tt \small \{ldynx,wonsikshin,nojunk\}@snu.ac.kr \hspace{.2cm} 
\{junsoolee,jeesookim\}@webtoonscorp.com}
 \\
{\small \url{https://ldynx.github.io/SAVE/}}
}

\newif\ifreview 
\newif\ifarxiv \newcommand{\arxiv}{\arxivtrue}
\newif\ifcamera 
\newif\ifrebuttal

\arxiv 

\makeatletter
\@namedef{ver@everyshi.sty}{}
\makeatother

\pdfoutput=1
\documentclass[10pt,twocolumn,letterpaper]{article}
\ifreview \usepackage[review]{cvpr} \fi
\ifarxiv \usepackage[pagenumbers]{cvpr} \fi
\ifrebuttal \usepackage[rebuttal]{cvpr} \fi
\ifcamera \usepackage{cvpr} \fi

\usepackage{graphicx}
\usepackage{amsmath}
\usepackage{amssymb}
\usepackage{booktabs}
\usepackage{multirow}
\usepackage[dvipsnames]{xcolor}
\usepackage[export]{adjustbox}

\usepackage{xr-hyper}

\makeatletter
\newcommand*{\addFileDependency}[1]{
  \typeout{(#1)}
  \@addtofilelist{#1}
  \IfFileExists{#1}{}{\typeout{No file #1.}}
}

\makeatother

\definecolor{cvprblue}{rgb}{0.21,0.49,0.74}
\usepackage[pagebackref,breaklinks,colorlinks,citecolor=cvprblue]{hyperref}

\begin{document}
\title{\paperTitle}
\author{\authorBlock}

\twocolumn[{%
\renewcommand\twocolumn[1][]{#1}
\maketitle
\begin{center}
    \centering
    \captionsetup{type=figure}
    \includegraphics[width=.82\textwidth]{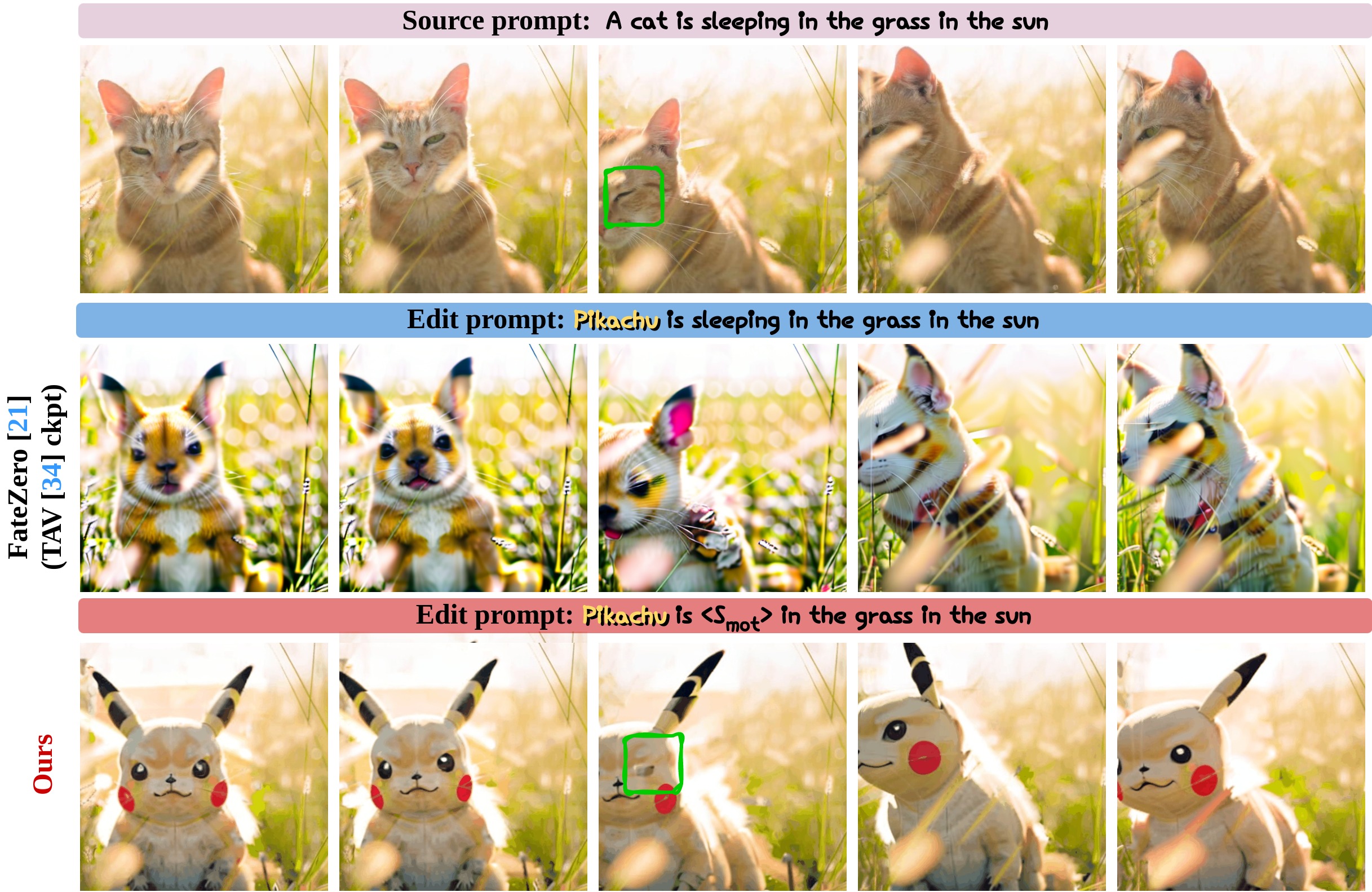}
    \captionof{figure}{\textbf{Diversification of a protagonist}. We present a single-shot video editing method that replaces the protagonist of a source video with the one described by the editing prompt maintaining the motion of the original protagonist.
    Different from previous works, our method is able to cope with diverse protagonists with substantial changes in their body structure.
    While other methods either fail to follow the editing prompt (second row) or generate a different motion (see Fig.~\ref{fig:qualitative}), our method achieves success in both challenges (third row).}
    \label{fig:introduction}
\end{center}
}]

\begin{abstract}
Driven by the upsurge progress in text-to-image (T2I) generation models, text-to-video (T2V) generation has experienced a significant advance as well. Accordingly, tasks such as modifying the object or changing the style in a video have been possible.
However, previous works usually work well on trivial and consistent shapes, and easily collapse on a difficult target that has a largely different body shape from the original one.
In this paper, we spot the bias problem in the existing video editing method that restricts the range of choices for the new protagonist and attempt to address this issue using the conventional image-level personalization method.
We adopt motion personalization that isolates the motion from a single source video and then modifies the protagonist accordingly.
To deal with the natural discrepancy between image and video, we propose a motion word with an inflated textual embedding to properly represent the motion in a source video.
We also regulate the motion word to attend to proper motion-related areas by introducing a novel pseudo optical flow, efficiently computed from the pre-calculated attention maps. Finally, we decouple the motion from the appearance of the source video with an additional pseudo word. Extensive experiments demonstrate the editing capability of our method, taking a step toward more diverse and extensive video editing.

\vspace{-3mm}
\end{abstract}
\vspace{-3mm}
\section{Introduction}
The remarkable advancements in text-to-image (T2I) generation models~\cite{ho2020denoising,song2020score,nichol2021glide,rombach2022high,ramesh2022hierarchical,saharia2022photorealistic} have prompted an increasing demand for the generation of imaginative scenes featuring user-supplied personalized concepts~\cite{gal2022image,ruiz2023dreambooth,kumari2023multi,tewel2023key,wei2023elite}. With these personalization methods, one can compose novel scenes with the desirable objects contained in various contexts \eg, pictures of \textit{the same} own dog traveling around the world. 
Some approaches~\cite{huang2023reversion,sohn2023styledrop} have expanded personalized concepts into a higher level beyond substantial objects: a relation between objects~\cite{huang2023reversion} (\eg, a dog \textit{shakes hands with} a cat \textit{side by side}) or even an image's style~\cite{sohn2023styledrop}.

On the other hand, there have been many video methods~\cite{ho2022imagen,hong2022cogvideo,singer2022make,wu2022nuwa} that adopt the weights of T2I models leveraging the extensive T2I prior knowledge from large-scale image datasets while inflating a model architecture to address temporal consistency. On top of this architecture, there are several existing methods~\cite{molad2023dreamix,esser2023structure,wu2022tune,liu2023video,zhang2023towards} proposed to edit the appearance and semantics of a given source video while preserving its geometry and dynamics. 

While they provide encouraging results in terms of frame consistency, the capability of understanding and reproducing a motion is confined within certain limitations. As shown in Fig.~\ref{fig:introduction} and Fig.~\ref{fig:qualitative}, state-of-the-art (SoTA) methods for video editing~\cite{wu2022tune,liu2023video,qi2023fatezero} have difficulty in generating a new protagonist whose body structure deviates significantly from that of the original protagonist to faithfully follow the motion in a source video. 

We have found that in the existing methods, cross-attention maps of a motion-related word (\eg `sleeping' in Fig.~\ref{fig:introduction}) are dispersed to regions that are not related to the motion as training proceeds. 
Therefore, learning and generating the specific motion in a source video become heavily dependent on temporal self-attention layers in a network architecture. However, the temporal self-attention layers only consider a temporal change in one latent pixel by its nature, disregarding the spatial relationship among pixels. This leads to a limited video editing capability when a new protagonist has a different shape and arrangement.

To solve this problem, we relieve the burden of the temporal self-attention and hand over the role to a more appropriate component, the word embedding vector to capture the motion. We reinterpret the protagonist editing task as a motion inversion problem and establish the following two goals: (1) a broaden personalized concepts expanded to \textbf{\textit{a motion}} in a source video and (2) the generation of a conceptualized motion with various contexts \ie protagonists. We introduce a new motion word ($S_{mot}$) that describes a specific motion performed by a protagonist in a source video. We have two advantages in utilizing $S_{mot}$ in editing a protagonist across a wider spectrum. First, at training, features of this motion word are injected in cross-attention mechanism based on a calculation of the attention map over a spatial axis. Therefore, the spatial characteristics of the learned motion can be fully explored during training.
Second, at inference, an embedding vector of motion word exchanges information with another embedding vector of the protagonist word (\eg `Pikachu' in Fig.~\ref{fig:introduction}) via the text encoder layer in the early stage. Then, the overall structure of a new protagonist in the motion can be determined from the start, allowing a natural movement in the edited video. The temporal self-attention layers, meanwhile, are able to concentrate more on a temporal change in each latent pixel. 

However, a pseudo-word~\cite{gal2022image,tewel2023key,wei2023elite,huang2023reversion} designed for image-level personalization leads to the discrepancy between image and video. Therefore, we expand the temporal axis of the textual embedding space that enables $S_{mot}$ to find a proper cross-attention map on moving areas. Moreover, to {let $S_{mot}$ make an effect on} a motion-related region and encourage effective motion learning, we introduce a novel \textit{pseudo} optical flow. From pre-calculated spatio-temporal attention maps, we track the semantically same pixels across frames and estimate the flows in a source video without requiring an extra optical flow model that is computationally expensive. Using this pseudo optical flow, we specify a motion-related region and better involve $S_{mot}$ in this area. During training, we also adopt an additional pseudo-word $S_{pro}$ representing the protagonist in a source video. Training $S_{mot}$ after registering $S_{pro}$ into the model's dictionary alleviates the entanglement between the motion and the protagonist. As shown in the last row of Fig.~\ref{fig:introduction} and Fig.~\ref{fig:qualitative}, given a single source video specialized to a specific motion, our method flexibly covers a broad range of the protagonist editing.

\section{Related Work}
\noindent
\textbf{Video Diffusion Models.} Leveraging the extensive prior knowledge of the image diffusion model has led to research endeavors~\cite{ho2022imagen, hong2022cogvideo, singer2022make, zhang2023show, guo2023animatediff} aimed at generating high-quality videos. 
In video generation, it is crucial to maintain consistency across generated frames--\textit{temporal consistency}.
As the image diffusion model has not learned information pertaining to temporality, these methods focus on injecting temporal information into a model architecture while retaining the existing wealth of spatial knowledge. 
Research also extends beyond video generation, delving into the realm of video editing and enabling straightforward modifications of videos provided by users~\cite{esser2023structure,molad2023dreamix, bar2022text2live, wu2022tune, liu2023video, qi2023fatezero, zhang2023towards}.
Recently, zero-shot editing approaches have been proposed~\cite{geyer2023tokenflow, zhang2023controlvideo} that alter the overall style of the video. 
The outcomes of these studies are closely tied to the structure of the input video with minimal changes to the structure of the output rather than directly modeling the motion information. A significant distinction in our work lies in the potential for video editing even when there is substantial alteration in the body structure of objects, setting it apart from the aforementioned studies. Concurrent to our work, there are methods~\cite{zhao2023motiondirector,wu2023lamp} to achieve generalization of the motion pattern from a given set of video clips. Our work focuses on customizing a unique, specific motion in a \textit{single} source video.

\noindent
\textbf{Personalization.} In diverse research endeavors, methodologies for generating \textit{specific concepts} beyond proficient creation have been studied in parallel. 
The most representative form is to generate a specific object a user gives. 
To this end, previous studies have investigated where and to what extent weights of the pre-trained model should be finetuned: a special token~\cite{gal2022image}, text encoder~\cite{wei2023elite}, keys of attention layers~\cite{tewel2023key}, and the whole model~\cite{ruiz2023dreambooth}.
There is research~\cite{avrahami2023break, xiao2023fastcomposer} exploring the combination of various concepts.
Moreover, research extends beyond the scope of objects such as style~\cite{sohn2023styledrop} or relationship~\cite{huang2023reversion}.
In this paper, we aim to integrate personalization with the video domain by endowing the capability of editing the protagonist while preserving the original motion information within a source video.
\section{Preliminaries}

\noindent
\textbf{Latent Diffusion Models (LDMs).} 
LDMs are diffusion models that recursively denoise an image latent code $z_t$ (backward process) into the previous timestep image latent code $z_{t-1}$ which is generated by repetitively adding noise to $z_0$ (forward process).
If needed, we can add a condition $\mathcal{C}$ during the backward process and the objective of LDMs can be defined as minimizing the following LDM loss:
\begin{align} \label{eq:ldm_loss}
    \mathcal{L}_{\text{ldm}} = \mathbb{E}_{z_0, \epsilon \sim \mathcal{N}(0, I), t \sim \text{Uniform}(1, T)} \left\| \epsilon - \epsilon_{\theta}(z_t, t, \mathcal{C}) \right\|_2^2,
\end{align}
where $z_t$ is the latent code of timestep t and $\epsilon$ is a random noise that the model $\epsilon_\theta(\cdot)$ should predict.

\noindent
\textbf{Text-to-Video Diffusion Models (T2V).}
To generate videos, the Video Diffusion model $\epsilon_{\theta}$ is structured with 3D UNet architecture~\cite{wu2022tune,blattmann2023align, liu2023video, guo2023animatediff, zhang2023controlvideo}.
Within the 3D UNet, there are layers dedicated to handling spatial and temporal information. The spatial layers are initialized from the weights of the image diffusion model, leveraging its extensive knowledge. We employed \textit{spatio-temporal} attention (ST-Attn) in the first block of the UNet, a method that considers the first frame along with the preceding frame when generating each frame. Meanwhile, \textit{temporal self-attention} (T-Attn) layers in the last block of the UNet aim to align frames by processing videos in a temporal dimension. These approaches aid in maintaining temporal consistency.
\section{Method}

We aim to propose a method that can edit a protagonist of a single source video while maintaining its motion. In doing so, we would like to introduce a way of enabling a broader range of choices for a new protagonist. We first discover the underlying causes of why the existing methods struggle to change a protagonist into a new object that is dissimilar in shape and arrangement (Sec.~\ref{subsec:location_bias}). Then, we introduce our method that utilizes a new motion word to transcend the limitation of the existing works (Sec.~\ref{subsec:expanded_text}) as well as an additional regularization term and a training strategy that guides the motion word to effectively learn the motion in a source video (Sec.~\ref{subsec:cross_attention_loss} and Sec.~\ref{subsec:pre-registration}).

\subsection{Location Bias of Motion-Related Words}
\label{subsec:location_bias}

\begin{figure}[t]
    \centering
  \includegraphics[width=1\linewidth]{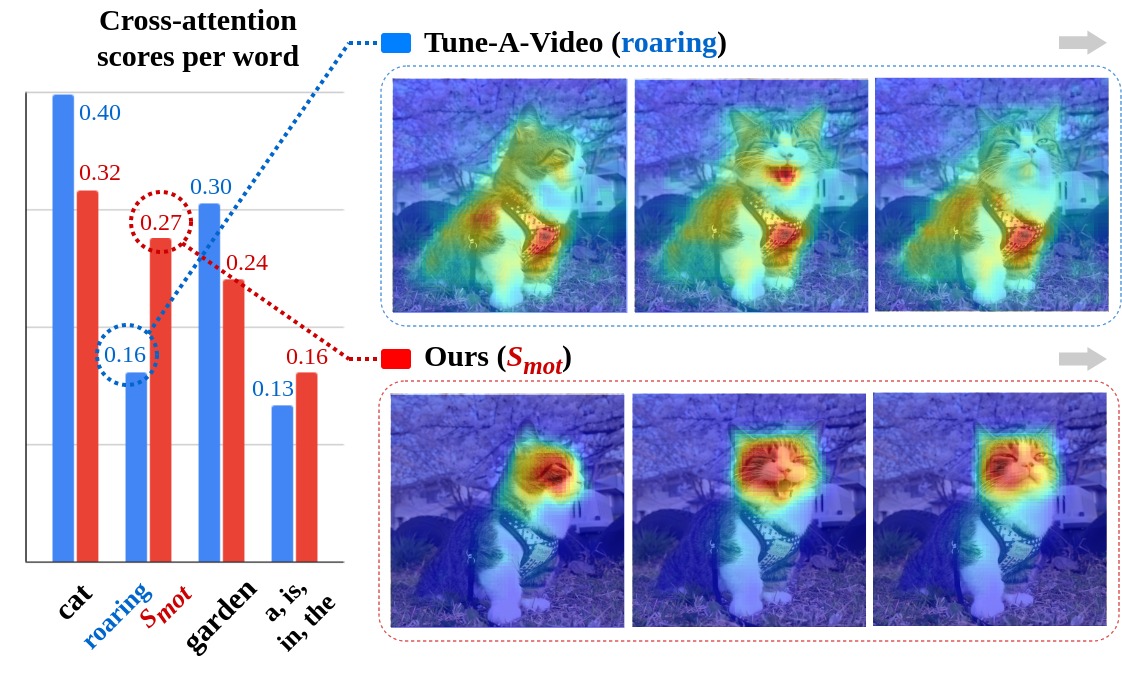}\\
  \vspace{-2mm}
  \caption{\textbf{Cross-attention scores and the attention maps of the motion-related word.} Using the existing method~\cite{wu2022tune} and ours, we compute the cross-attention scores of each semantic token and visualize regions to which the motion-related word (`roaring' and $S_{mot}$) attends. Little information about `roaring' is used in \cite{wu2022tune} compared to the nouns (cat and garden) and it attends to an inaccurate region due to location bias. Meanwhile, our method actively utilizes $S_{mot}$ which provides more accurate information about the motion attending to the proper facial regions.}
  \label{fig:image_bias}
  \vspace{-2mm}
\end{figure}

\begin{figure*}[t]
    \centering
  \includegraphics[width=1\linewidth]{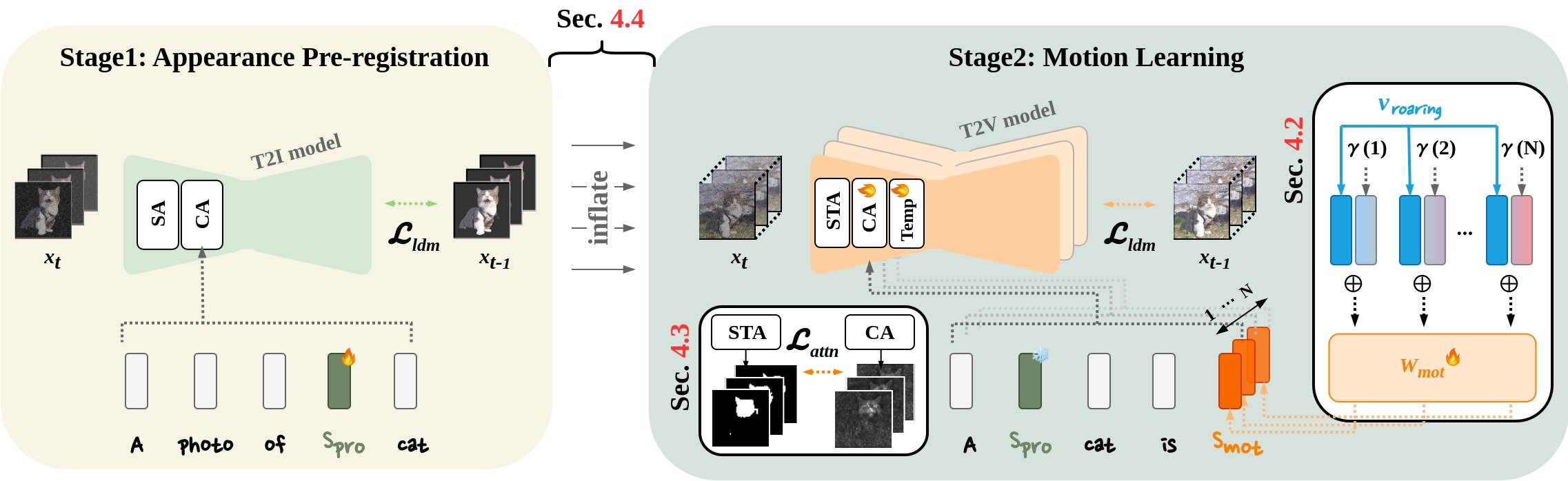}\\
  \caption{\textbf{The overall training pipeline of our method.} Using expanded text embeddings of $S_{mot}$, we optimize $W_{mot}$ that maps embeddings of the original motion word (`roaring`) to a specific motion in a source video. Under cross-attention regularization, $S_{mot}$ is optimized to primarily focus on the moving area while pre-registered $S_{pro}$ disentangles the appearance from the motion. This dual approach facilitates $S_{mot}$ in effectively learning the motion.}
  \label{fig:pipeline}
  \vspace{-2mm}
\end{figure*}

As shown in Fig.~\ref{fig:introduction} and Fig.~\ref{fig:qualitative}, the existing methods fail to produce a new, differently structured protagonist appropriately following the motion in the source video. 
To identify the causes, we start by investigating a bias in cross-attention maps of motion-related words. We have found that those words often highlight specific parts of the protagonist apart from motion-related areas.
As shown in the first row of Fig.~\ref{fig:image_bias}, cross-attention maps of motion-related words (\eg `roaring') scatter over the neck of a cat. We define this phenomenon as \textit{location bias}.

Why does this bias appear especially in motion-related words? The pretrained text encoder is usually trained on large-scale text-image datasets and this makes the encoder to embed motion-related words into static visual appearance (\eg `roaring' commonly generates images with a mouth open wide to the object's neck). However, in a video, motion proceeds in the stream of time where motion-location relation exists only in several specific frames (\eg `roaring' video also includes frames where the mouth is closed).
Naturally, the cross-attention map of those words lies on a limited area of an image.

As location biases produce inaccurate features injecting textual information in an incorrect position, the model heavily relies on T-Attn layers instead of utilizing a motion-related word. The left graph in Fig.~\ref{fig:image_bias} shows that textual information participates little in the motion generation process (only 16\% of total attention is on the verb term). In Fig.~\ref{fig:analysis}, we also show that the model largely depends on T-Attn layers where the motion cannot be reconstructed without training T-Attn layers. 
Meanwhile, T-Attn layers {remain irrelevant to the} spatial axis by its nature. 
When the input has $B$ batch size, $N$ sequence length, and $H \times W$ spatial dimension, they treat $B \cdot H \cdot W$ encoded vectors independently exchanging information only among $N$ features in the same latent pixel. For the $k$-th pixel on a latent code from the $i$-th frame $z^{i,k}$ as the query, T-Attn is calculated with $z^{j,k}, j \in \{1, \cdots, N\}/i$ as key.
\begin{align}
Q = W^Qz^{i,k}, K=W^Kz^{j,k}, V=W^Vz^{j,k}
\end{align}
where $W^Q,W^K $ and $W^V$ are projection matrices.
Therefore, when T-Attn layers encounter a protagonist with a new body structure, they cannot reproduce a proper motion on the protagonist. In Fig.~\ref{fig:qualitative}-left, for example, a newly generated dog cannot achieve a learned motion around the head and the mouth since the original protagonist (a cat) and edited protagonist (a dog) have a dissimilar facial arrangement and mouth shape.

\subsection{Expanded Text Embeddings with Time} \label{subsec:expanded_text}
As our goal is to put a specific motion on various types of protagonists, we focus on the approach to revitalizing a motion-related word by reducing location biases. To this end, we expand the textual embedding space of a motion word to represent a time flow in videos rather than a frozen moment in images: we add a temporal axis to an embedding space of our new motion word ($S_{mot}$) and let $S_{mot}$ inject its information into a proper region in \textit{each frame}.

To formulate embedding vectors of $S_{mot}$, we have two separate components to take on different roles. The first component utilizes prior knowledge of the T2I model and represents the static appearance of the motion, conveying its information to the second component. The second component learns the residual motion for each frame and encodes the overall motion in the video. The embedding vectors of $S_{mot}$ for $N$ video frames can be gained as follows:
\begin{align} \label{eq:v_mot}
    v_{mot}^{i} = W_{mot}\ (v_{b} \oplus \gamma(i)),\quad i \in \{1, \cdots, N\}
\end{align}
where \textbf{$v_{b}$} is a textual embedding of the original motion-related word in a source prompt and \textbf{$W_{mot}$} is learnable linear layers. We denote concatenation operation and positional encoding as $\oplus$ and $\gamma(\cdot)$ respectively. $v_{mot}^{i}$ is then treated like other embedding vectors and passed through a text encoder. 

Contrary to conditioning a single text embedding to $N$ latent codes in the existing text-to-video methods, we provide $N$ text embeddings, $v_{mot}^i, \ i \in \{1, \cdots, N\}$ to each latent code of the corresponding frame. As shown in the second row of Fig.~\ref{fig:image_bias}, with consideration for a time flow, our $S_{mot}$ renders more accurate features via cross-attention whose attention maps flexibly attend to moving areas of each frame. Note that in spite of an increasing number of video frames, the number of learnable parameters stays fixed, enabling video editing in various lengths.

\subsection{Motion Aware Cross-attention Loss} \label{subsec:cross_attention_loss}

\begin{figure}[t]
    \centering
  \includegraphics[width=1\linewidth]{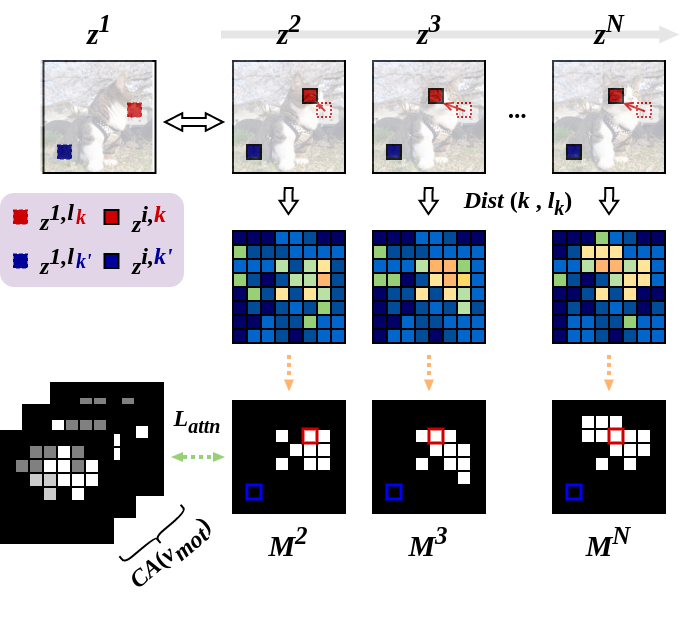}\\
  \vspace{-4mm}
  \caption{\textbf{Cross-attention regularization on $S_{mot}$.} We assume that a pixel $z^{i,k}$ in the moving areas becomes distant from a pixel $z^{j, l_{k}}$ that has the maximum attention score ($l_k$ indicates $l^{j,k}_{max}$) while a pixel $z^{i,k'}$ in the static areas stays close to $z^{j, l_{k'}}$. Therefore, we calculate the distance between $z^{i, k}$ and $z^{j, l_k}$ for all $k$ and generate motion masks $M_2, \cdots M_N$. Note that $j \in \{1, i-1\}$ and we illustrate the process when $j=1$ for simplicity.}
  \label{fig:motion_map}
  \vspace{-2mm}
\end{figure}

As Perfusion~\cite{tewel2023key} points out, a cross-attention map of a newly learned concept is {likely to pervade} the whole scene, easily overfitting to the background. While a video is composed of various {dynamics} \eg camera moving or background movement, {we want $S_{mot}$ to learn specifically the motion of the protagonist.} Inspired by \cite{avrahami2023break,xiao2023fastcomposer}, we constrain a cross-attention map of $S_{mot}$ to focus exclusively on the protagonist. However, forcing $S_{mot}$ to pay attention on the entire protagonist obscures what $S_{mot}$ needs to learn \ie the motion of the protagonist.

Therefore, we propose a motion-aware cross attention loss enabling $S_{mot}$ to focus on the movement of the protagonist. We also introduce a novel pseudo optical flow to better represent the moving area without using the costly optical flow models. We first collect pre-calculated ST-Attn maps from specific decoder layers. In ST-Attn mechanism, the $i$-th attention map is computed by {using the $i$-th frame as the query and the first and the preceding frame} as keys.

We refer a latent code from the $i$-th frame as $z^i$. Attention scores $SA(z^i, z^j)$ then represent a similarity between $z^i$ and $z^j, j \in \{1, i - 1\}$. Our intuition lies on that if the $k$-th pixel of the $i$-th frame $z^{i,k}$ and the $l$-th pixel {of the $j$-th frame} $z^{j,l}$ have a high score, then they ten to be the same semantic point at different frames, $i$ and $j$. Therefore, by tracking down spatial locations of these similar points across frames, we can estimate a temporal flow of each pixel in the video.

To find the points that are likely to be the same, we store a spatial location $l^{j, k}_{max} \in [0, h-1] \times [0, w-1]$ of $z^j \in \mathbb{R}^{c \times h \times w}$ where the maximum score with {$z^{i,k}$} exists.
Here, $c$, $h$, and $w$ indicate the channel size and spatial dimensions of $z^j$ respectively. Then, we calculate a distance between $l^{j, k}_{max}$ and {$z^{i,k}$}. When two locations are close to each other, the object in {$z^{i,k}$} can be regarded as mostly stationary until the $i$-th frame. On the other hand, if $l^{j, k}_{max}$ is far from {$z^{i,k}$}, then the object corresponding to the {$z^{i,k}$} is highly likely to be moving in the video. 

We go through the same process for each {pixel $k$ of the $i$-th frame} and produce a map with the estimated distance which amounts to the moving distance of a pixel from the $j$-th frame. Eventually we extract masks of motion-related area for each frame $i \in \{2, \cdots, N\}$ {by retaining only pixels with a large moving distance}. We denote the masks as $M = \{M^2, \cdots, M^N\}$. Utilizing these motion masks, we add the following regularization term to encourage cross-attention map of $S_{mot}$ to follow $M$:
\begin{align} \label{eq:attn_loss}
\vspace{-0.3mm}
\mathcal{L}_{\text{attn}} = \frac{1}{N-1} \sum_{i=2}^N \Vert CA(z^i, v_{mot}^i) - M^i \Vert_2^2
\vspace{-0.3mm}
\end{align}
We illustrate the overall regulating process in Fig.~\ref{fig:motion_map}. 
We also include more details in the supplementary materials

Adopting Eq.~\ref{eq:ldm_loss} and Eq.~\ref{eq:attn_loss}, the overall optimization objective is defined as:
\begin{align}
\vspace{-0.3mm}
    \mathcal{L} = \mathcal{L}_{\text{ldm}} + \lambda_{\text{attn}}\mathcal{L}_{\text{attn}}
\vspace{-0.3mm}
\end{align}
where $\lambda_{\text{attn}}$ is a hyperparameter balancing between a reconstructive ability and a motion focusing.

\subsection{Appearance Pre-registration Strategy} \label{subsec:pre-registration}
We assume that we have a single video in hand containing a particular motion we want to quote. However, the motion and the protagonist get easily entangled. To resolve this problem, we propose a two-stage training strategy to untangle the two properties.

We newly define a pseudo-word $S_{pro}$ that represents the appearance and texture features of the protagonist. At the first stage, we find a text embedding of $S_{pro}$, \textbf{$v_{pro}$}, in the textual embedding space without a temporal axis, disregarding the motion. With a pretrained T2I model, \textbf{$v_{pro}$} is optimized with the LDM loss as in Eq.~\ref{eq:ldm_loss} considering video frames as batch of images. At the second stage, we inflate the T2I model to a T2V model and optimize \textbf{$W_{mot}$} and \textbf{$v_{b}$} in Eq.~\ref{eq:v_mot}. As the protagonist and its appearances are already registered in the text encoder, $v_{mot}$ can be effectively learned using disentangled motion information for the video. 

\begin{figure*}[t]
    \centering
  \includegraphics[width=1\linewidth]{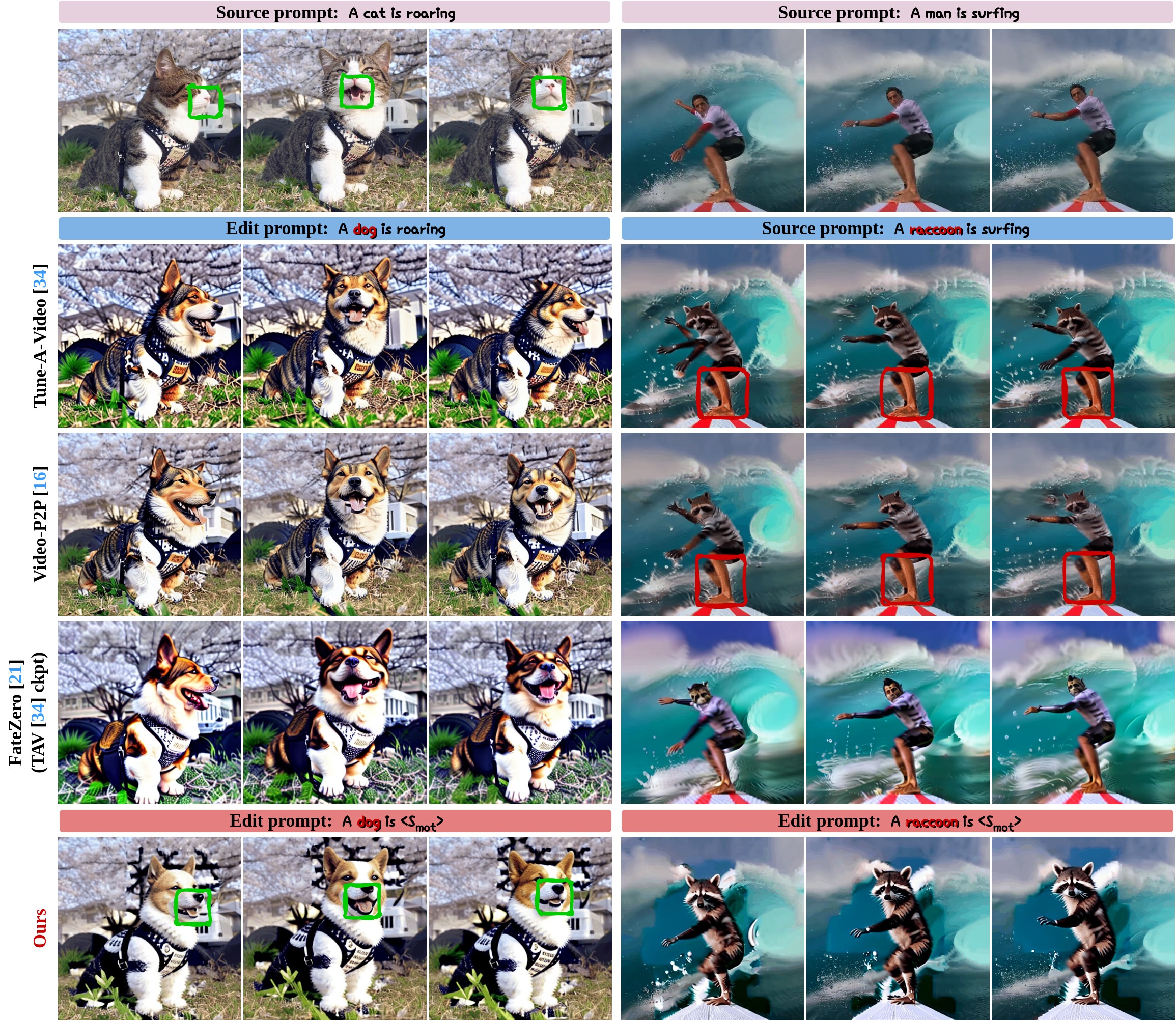}\\
  \caption{\textbf{Protagonist editing results comparing our method with other baselines.} Our method successfully reproduces the motion in a source video identifying $S_{mot}$ as both the head and the mouth movements (left column) while editing a protagonist faithfully with a natural video of a raccoon \textit{doing $S_{mot}$} (right column). Meanwhile, other baselines either generate an inaccurate motion \eg a dog opening its mouth across all frames, or fail to combine a new protagonist with the motion resulting in incomplete editing.}
  \label{fig:qualitative}
  \vspace{-2mm}
\end{figure*}

\section{Experiments}

\textbf{Dataset.} We evaluate our method and baselines on videos collected from DAVIS dataset~\cite{perazzi2016benchmark} and YouTube~\cite{liu2023video} following previous works~\cite{wu2022tune,liu2023video,bar2022text2live}. Each video consists of 8--32 frames at the resolution of 512 $\times$ 512. As one of the important aspects to evaluate is the editing ability of a protagonist with large structural changes, we additionally provide object-changed prompts for hard cases (\eg changing a cat in a source video to Pikachu). Ultimately, we composed 48 pairs of videos and text prompts to evaluate. We also conduct additional experiments on the open-sourced benchmark released by the LOVEU-TGVE competition at CVPR 2023~\cite{wu2023cvpr}. More details as well as qualitative results on LOVEU dataset are in the supplementary materials.

\noindent
\textbf{Baselines.} We compare our method with the state-of-the-art video editing and generation approaches. \textit{(1) Tune-A-Video (TAV)}~\cite{wu2022tune} is the conventional video editing method that fine-tunes the inflated T2I model on a given source video. \textit{(2) Video-P2P}~\cite{liu2023video} improves upon TAV applying Prompt-to-Prompt~\cite{hertz2022prompt} and Null-text Inversion~\cite{mokady2023null} to the T2V model. \textit{(3) Fate-Zero}~\cite{qi2023fatezero} proposes to blend the attention maps stored during inversion. Following the one-shot tuning version, we adopt the TAV weights pretrained on the source video when evaluating Fate-Zero.

\begin{table}[t]
  \centering
  \footnotesize
    \caption{\textbf{Quantitative comparison on videos collected from DAVIS dataset~\cite{perazzi2016benchmark,wu2022tune,bar2022text2live} and YouTube~\cite{liu2023video}.}}
    \vspace{-2mm}
  \begin{tabular}{l ccc}
    \toprule
    Method & \multicolumn{3}{c}{Automated Metrics}\\
    \cmidrule(lr){2-4}
    & 
    {\parbox{1.5cm}{\centering CLIP-Text $\uparrow$}} & 
    {\parbox{1.5cm}{\centering CLIP-Img $\uparrow$}} &
    {\parbox{1.5cm}{\centering Flow sim. $\uparrow$}} \\
    \midrule
    Tune-A-Video~\cite{wu2022tune} & 25.84 & 93.37 & 55.61 \\  
    Video-P2P~\cite{liu2023video} & 25.27 & 93.89 & 64.59 \\  
    Fate-Zero~\cite{qi2023fatezero} & 24.14 & 94.05 & \textbf{80.92} \\ 
    \midrule
    Ours & \textbf{25.99} & \textbf{94.21} & 79.10 \\  
    \bottomrule
  \end{tabular}
  \label{tab:quatitative}
  \vspace{-2mm}
\end{table}

\begin{figure}[t]
    \centering
  \includegraphics[width=.8\linewidth]{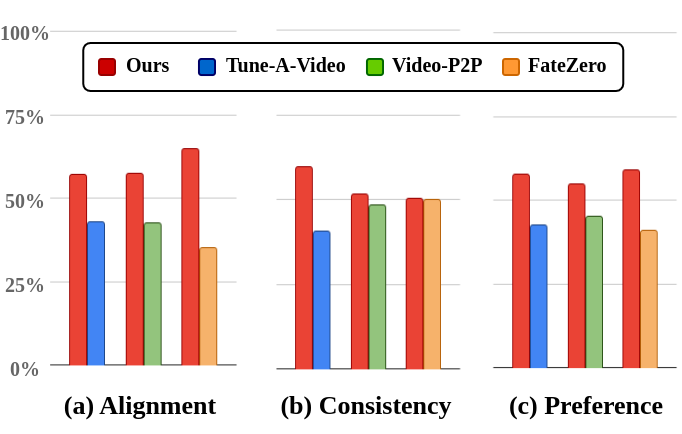}\\
  \vspace{-2mm}
  \caption{\textbf{User study results.} Our method is preferred over other baseline methods across all evaluation criteria.}
  \label{fig:user_study}
  \vspace{-2mm}
\end{figure}

\noindent
\textbf{Metrics.} In line with previous works~\cite{wu2022tune,esser2023structure,wang2023zero}, we evaluate the baselines using the pretrained CLIP~\cite{hessel2021clipscore} model as follows: \textit{(1) CLIP-Text similarity} is the average CLIP score between frames of generated video and the corresponding editing prompt, representing a textual alignment of the outputs. \textit{(2) CLIP-Image similarity} computes cosine similarity between the CLIP image embeddings of pairs of video frames, representing frame consistency. To score faithfulness to the motion of the source video, we measure \textit{(3) Flow similarity} that computes cosine similarity between optical flows of the source and the edited video using the estimation model~\cite{teed2020raft}. We further evaluate the methods through five human raters for each example conducted with Amazon Mechanical Turk. The following three questions were asked. \textit{(1) Textual alignment}: ``Which video better matches the text?'' \textit{(2) Consistency}: ``Which video has higher consistency?'' \textit{(3) Preference}: ``From the perspective of video editing, which video do you prefer?''

\subsection{Comparison with the Baselines} \label{subsec:comparison}

\noindent
\textbf{Qualitative results.} Fig.~\ref{fig:qualitative} illustrates qualitative comparisons between our method and baselines. As our method (bottom row) {effectively learns the motion of the original protagonist,}
it generates a new protagonist that reproduces the motion in the source video seamlessly despite having a significantly different structure from that of the original one. For example, as shown in Fig.~\ref{fig:qualitative}-left, our method ables to grasp the mouth movements as well as head-turning movement in the source video and successfully associates those movements to $S_{mot}$. Meanwhile, other baselines commonly generate a dog in the cat's silhouette from the source video and miss the mouth movements. TAV~\cite{wu2022tune} also produces an inaccurate head motion in the third frame.
Our method effectively reflects the editing prompts compared to other baselines as shown in the right column of Fig.~\ref{fig:qualitative}. Baseline methods are unable to overcome the discrepancy between `man' and `raccoon' in the structure and generate a raccoon video where certain segments maintain the man's appearance. On the other hand, our method disentangles the appearance and the motion with separate $S_{pro}$ and $S_{mot}$ and renders a new raccoon doing $S_{mot}$ from the text encoder from the start, successfully applying the motion features to a new protagonist. We provide additional interesting results in the supplementary material.

\begin{figure}[t]
    \centering
  \includegraphics[width=1\linewidth]{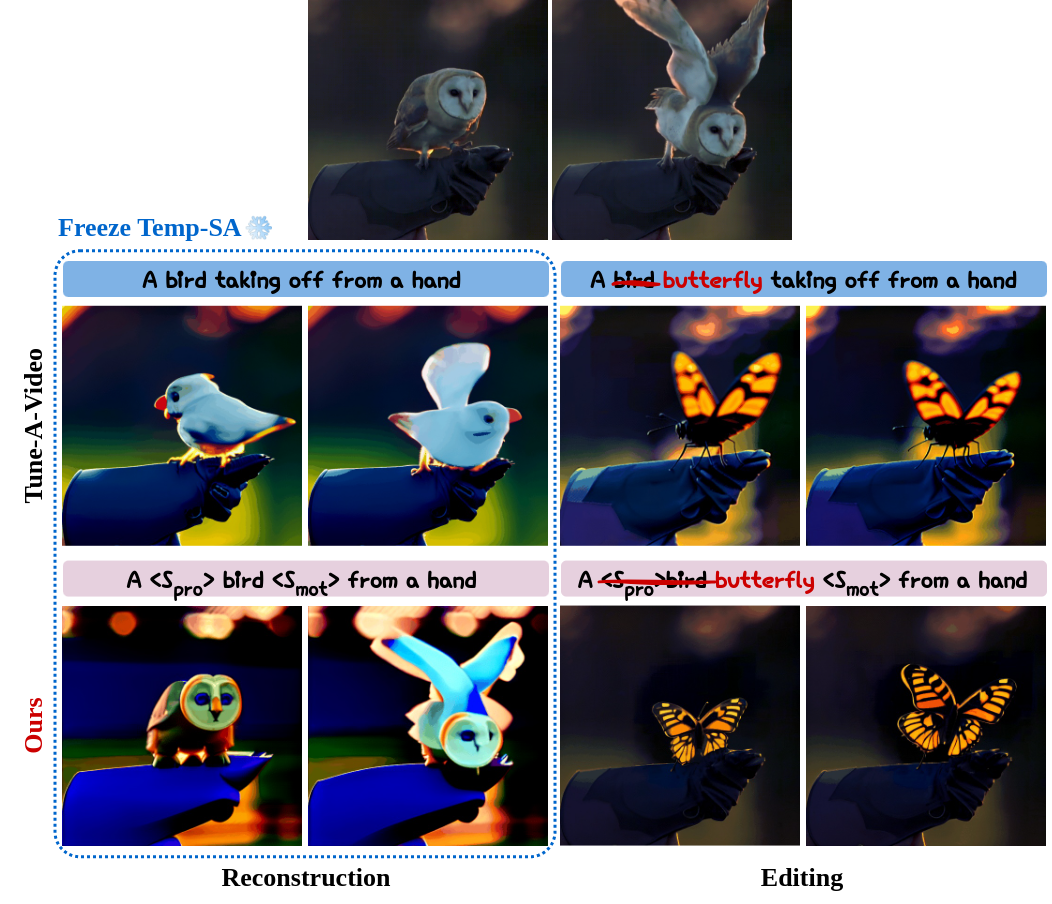}\\
  \vspace{-2mm}
  \caption{\textbf{Analysis on motion learning and additional editing results.} As $S_{mot}$ effectively learns the motion in a source video, our method still reproduces a proper motion with frozen T-Attn layers, resulting the more flexible editing.}
  \label{fig:analysis}
  \vspace{-2mm}
\end{figure}

\noindent
\textbf{Quantitative results.} In Tab.~\ref{tab:quatitative} and Fig.~\ref{fig:user_study}, our method exhibits the highest ability in the metrics of textual alignment, temporal consistency, and user preferences. As our method closely associates the learned motion with a new protagonist to edit, it successfully generates a natural video that is faithfully aligned with an editing prompt. Meanwhile, Fate-Zero~\cite{qi2023fatezero} shows the highest flow similarity and frame consistency on par with our method. As shown in Fig.~\ref{fig:qualitative} and the supplementary material, when an editing prompt requires a large structural change for a new protagonist, Fate-Zero shows a tendency to adhere closely to the source video. This leads to low CLIP-Text scores in Tab.~\ref{tab:quatitative} and less voted Alignment in Fig.~\ref{fig:user_study}, while the edited video still achieves a highly similar optical flow to that of the source video and a high frame consistency. On the other hand, our method attains high scores in \textit{both} text alignment and flow similarity \& consistency demonstrating a general editing ability.

\noindent
\textbf{Analysis on motion learning.} To demonstrate that incorporating $S_{mot}$ actually alleviates the burden on the T-Attn layers in learning the motion, we conduct the following experiment: we freeze the T-Attn layers in TAV and our method respectively when training the networks. After training, we reconstruct the source video using the motion-related words (`taking off' and $S_{mot}$). As shown in the left two columns in Fig.~\ref{fig:analysis}, TAV cannot properly reproduce a motion in the source video heavily depending on the T-Attn layers in regard of learning the motion. On the other hand, our method is able to generate the accurate motion in the source video by using $S_{mot}$. The right two columns in Fig.~\ref{fig:analysis} indicate editing results from each fully-trained method. Our method successfully renders a new protagonist undergoing structural modifications (\eg editing a bird to a butterfly) as $S_{mot}$ actively exploits the spatial information of the motion.

\begin{figure}[t]
    \centering
  \includegraphics[width=1\linewidth]{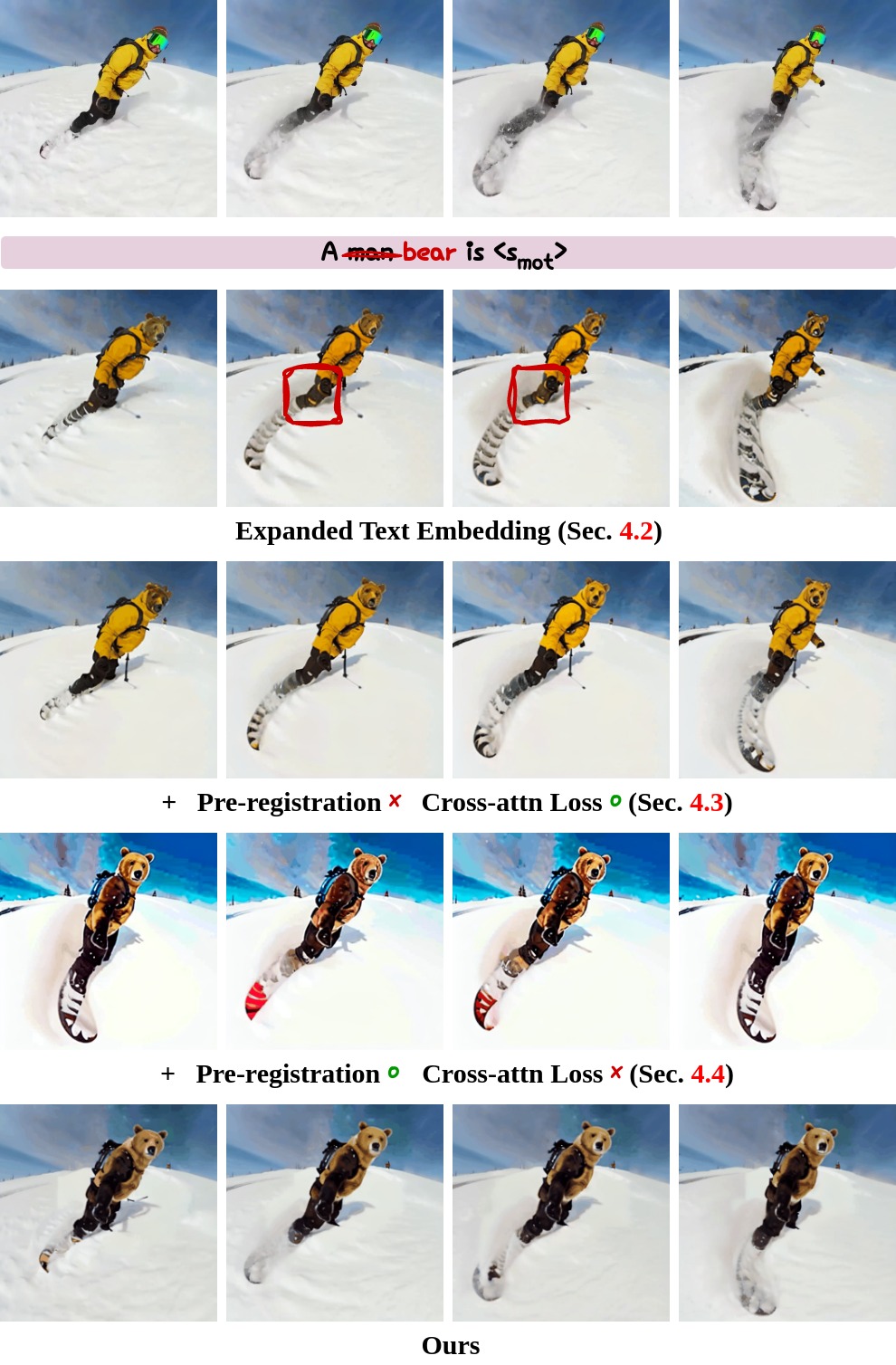}\\
  \vspace{-3mm}
  \caption{\textbf{The impact of each component.} By the temporally expanded text embedding, $S_{mot}$, our method is able to learn the motion across frames. Cross-attention regularization and pre-registration of the protagonist word further enhance an ability of $S_{mot}$ to understand the motion. }
  \label{fig:ablation}
\vspace{-3mm}
\end{figure}

\subsection{Ablation Studies} \label{subsec:ablation}

We isolate each component in our method and verify the effect respectively. As shown in the second row in Fig.~\ref{fig:ablation}, with temporally expanded text embedding $v_{mot}$, a new protagonist in the edited video well follows the overall pose of the original protagonist in the source video. However, the new protagonist exhibits an awkward leg appearance while the movement is also slightly different. By adopting cross-attention regularization, as shown in the third row, a generated bear retains more accurate movements in the source video. Meanwhile, pre-registration of $S_{pro}$ effectively decouples the motion from the appearance and generates a new protagonist faithfully as shown in the fourth row. The last row indicates the results of our method which disentangles the motion from the appearance in a source video and effectively digests the information on the motion. We refer to supplementary materials for ablations on other examples.

\begin{figure}[t]
    \centering
  \includegraphics[width=1\linewidth]{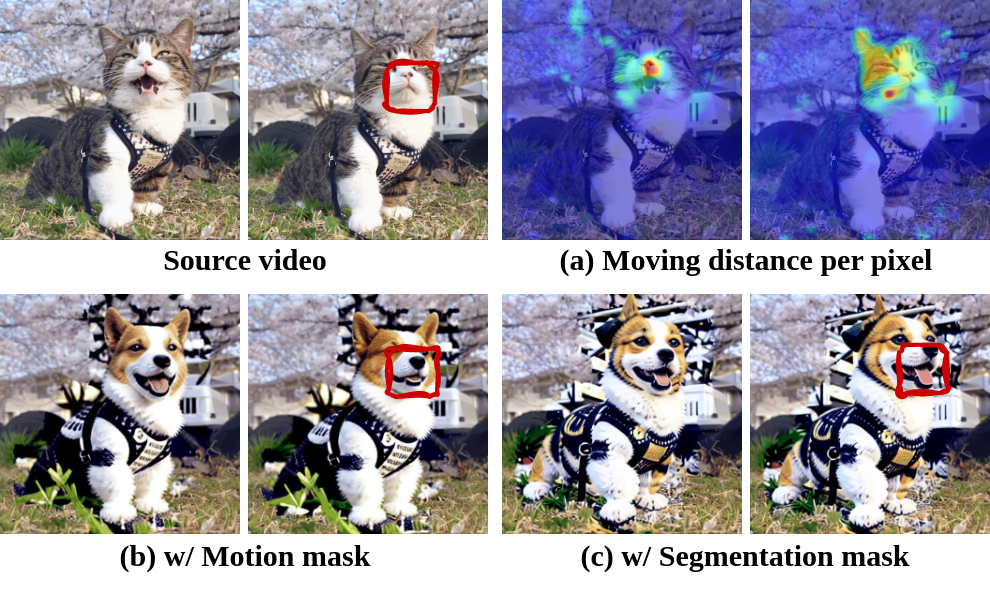}\\
  \vspace{-3mm}
  \caption{\textbf{Ablation on motion masks $M$.} Encouraging $S_{mot}$ to focus on motion masks guides $S_{mot}$ to learn the motion more effectively than using segmentation masks alone.}
  \label{fig:ablation_motion_map}
  \vspace{-2mm}
\end{figure}

We also investigate the efficiency of our motion masks $M$ introduced in Sec.~\ref{subsec:cross_attention_loss}. Fig.~\ref{fig:ablation_motion_map} illustrates an estimated moving distance of each pixel and our results using the motion masks $M$ in (a) and (b) respectively while comparing the results using the object segmentation masks~\cite{kirillov2023segment} instead in (c). Narrowing down the moving area with these motion masks effectively guides $S_{mot}$ to focus on the motion itself.

\section{Conclusion}
In this paper, we propose a new method to diversify a protagonist that reproduces the motion in a source video. We first reveal a location bias in the existing methods that hinders flexible editing. To resolve this problem, we introduce a motion word which encompasses temporal relationships among frames. We also adopt a couple of approaches to effectively train the motion word focusing on a target motion. Our method paves the way for broader editing, enriching the video editing task.

\noindent \textbf{Limitation \& Future work.} In this paper, we restrict a motion to the movement of a protagonist in a source video separate from background and camera movements. We also found that our method struggles to learn the motion of multiple protagonists as can be found in the failure cases in the supplementary materials. Future works to expand into broader movements will be an interesting topic.

{\small
\bibliographystyle{ieeenat_fullname}
\bibliography{main}
}


\end{document}